# A Large Language Models for Multi-Class and Multi-Label Detection of Drug Use and Overdose Symptoms on Social Media

Muhammad Ahmad[1,*], Fida Ullah[1], Muhammad Usman[1], Umyh Habiab[2], ldar Batyrshin[1] and Grigori Sidorov[1]

[1]Centro de Investigación en Computación, Instituto Politécnico Nacional (CIC-PN), Mexico City 07738, Mexico

[2] Independent Researcher

Correspondence Author: mahmad2024@cic.ipn.mx

## Abstract

Drug overdose is a serious health issue worldwide, leading the cause of serious medical complications and fatalities. Drug abuse particularly containing opioids, painkillers, and psychiatric medications has been identified as a key factor in the growing prevalence of addiction rate, increased criminal activity, and social instability. Traditional research approaches, such as clinical trials and surveys, often encounter methodological challenges such as insufficient sample sizes and response biases. However, social media has become a valuable platform where peoples share their self-reported experience including struggles, overdose symptoms, and recovery strategies. This study utilizes large-scale social media discourse to explore overdose patterns, identify regularly used substances and its overdose symptoms, and evaluate treatment approaches. The objective of this study is to utilize the social media textual data to train and test an AI-driven natural language processing (NLP) tool to analyze self-reported experience on substance abuse, and its overdose symptoms. This approach provides real-time insights into substance abuse trends, supporting public health surveillance, early intervention, and personalized care strategies. We employed hybrid data annotation by utilizing Large Language Models (LLMs) along with human annotators to annotate the dataset and explored various learning approaches for our task, including traditional machine learning models including TF-IDF, neural networks using pre-trained word embeddings such as FastText and GloVe, and state-of-the-art advance language based models with advanced contextual embeddings. These methods utilized natural language processing (NLP) techniques to develop an advanced automated tool that synthesizes these insights and uncovers hidden patterns in the textual data. The dataset was structured into two key components: (1) classification of multi class drugs used, (2) identification of multi-label overdose symptoms. The proposed AI-driven framework achieved 98% accuracy in multi-class classification and 97% in multi-label classification, surpassing baseline models with significant performance improvements. In the multi-class setting, it outperformed the best baseline model, Decision Tree (90%), by 8%, while in the multi-label setting, it exceeded the performance of Logistic Regression (92%) by 5%. The model demonstrated robust performance in accurately identifying drug types, overdose symptoms, and treatment methods from user-generated content. Statistical validation confirmed the reliability and effectiveness of our approach in detecting clinically relevant self-medicated and prescribed treatments for opioid withdrawal symptoms.

.

**Keywords:** Drug Overdose, Social Media Analysis, Large language model, GPT, Natural Language Processing (NLP), Multi-Label Classification, Public Health Surveillance, Machine learning, BERT, Deep learning.

## 1. Introduction

Drug overdose occurs when someone takes an excessive amount of legal or illegal substances, leading to serious health problems or even death. It has become a serious global health crisis [1–4], affecting millions of people and placing massive pressure on healthcare systems, economies, and families. The misuse of drugs [5–10]—including painkillers, sleeping pills, medications for mental health, and substances such as Alcohol, Cocaine, Ecstasy, Fentanyl, Heroin, Ketamine, Lysergic Acid Diethylamide (LSD), Methamphetamine, and Opioids—can lead to addiction, mental health disorders, increased crime rates, and social instability. Drug overdose continues to pose significant public health challenges worldwide, particularly in the United States [35]. In recent years, the U.S. government has significantly increased its budget to combat the overdose epidemic. For Fiscal Year (FY) 2024, President Biden proposed a historic investment of $46.1 billion for National Drug Control Program agencies, reflecting a $5.0 billion increase from the FY 2022 request and a $2.3 billion increase over the FY 2023 enacted level [34].

To tackle the issue of drug overdose, a global commitment to education, rehabilitation, and stronger policies are required to prevent abuse and improve treatment options. Traditional approaches to studying drug abuse [20-22] and overdose effects have relied on clinical trials, surveys, and self-reported data, which often suffer from limitations such as small sample sizes, response biases, and ethical constraints. However, with the growing influence of digital platforms, particularly social media, such as Twitter, Facebook, YouTube, and Reddit have emerged as a transformative platform, especially in the domains of healthcare [14–16] and mental health [17–19]. Many individuals also turn to these platforms to share their self-reported experiences with illicit [29] and prescribed drug abuse [11–13], its overdose symptoms, and recovery journeys. The fear of judgment or stigma often prevents people from seeking professional help, making them more comfortable opening up within online communities. These digital spaces offer a sense of anonymity and understanding, where individuals find support from others who are facing similar challenges. As a result, platforms such as Reddit have become essential platform for discussing personal experiences with substance abuse, its overdose symptoms, and harm reduction strategies. In Reddit there are subreddit in which Individuals often share how they became addicted to substances and describe the intense physical and emotional toll of overdose.

This growing body of user-generated content highlights the severity of the public health crisis caused by substance abuse and overdose, which continues to claim hundreds of thousands of lives every year [28, 33]. The widespread use of platforms like Reddit, Twitter, Telegram, and Snapchat allows users to easily access content related to drug use, including testimonials, usage guides, and sourcing tips. Additionally, online drug markets exploit social media to attract buyers, often presenting drugs as trendy, therapeutic, or harmless. This convergence of demand, visibility, and accessibility has amplified the risks of drug abuse and complicated law enforcement efforts to monitor illegal sales and prevent exposure to high-risk substances. Commonly misused substances—such as Alcohol, Cocaine, Ecstasy, Fentanyl, Heroin, Ketamine, Lysergic Acid Diethylamide (LSD), and Methamphetamine—pose serious health risks, each with distinct overdose profiles. Potent opioids like Fentanyl and Heroin are particularly deadly due to their high toxicity, while stimulants such as Cocaine and Methamphetamine are linked to dangerous cardiovascular and neurological effects.

To better understand and respond to this crisis, Natural Language Processing (NLP) offers powerful tools for analyzing social media conversations about drug use and overdose. Techniques including sentiment analysis [31], named entity recognition (NER) [30] [48], and topic modeling [32] can uncover patterns in how people describe their symptoms, experiences, and emotional states. These methods help identify early warning signs of overdose—such as drowsiness, confusion, slowed breathing, or unconsciousness—and track trends in public discourse. Beyond overdose detection, NLP has also been widely applied in other domains such as hate speech detection [48], misinformation identification, and hope speech detection [49], helping to monitor harmful content, promote safety, and support mental health in online spaces. Integrating these diverse NLP applications can offer a more comprehensive understanding of online health-related behaviors and societal risks.

This study aims to develop an automated advanced language-based approach that can quickly and accurately analyze social media posts related to specific drug use and overdose effects. It focuses on two main tasks: (1) identifying the drug consumed by the user and (2) identifying the associated overdose symptoms. These symptoms represent a multi-label classification problem, as individuals often experience multiple effects simultaneously. For instance, a person may report extreme agitation, chest pain at the same time or describe symptoms such as blue lips along with slow or shallow breathing. These combinations are not only common across various substances such as heroin, cocaine, and others, but also clinically meaningful, making multi-label detection essential for effective analysis.

To achieve this objective, we sourced Reddit posts related to eight high-risk drugs including Alcohol, Cocaine, Ecstasy, Fentanyl, Heroin, Ketamine, Lysergic Acid Diethylamide (LSD), and Methamphetamine. These drugs are known for their extreme physiological and psychological effects, with many linked to life-threatening overdoses. After the collection of dataset, we utilized hybrid data annotation schema by employing a Large Language Model (LLM) along with human annotators to annotate the corpus and applied various learning approaches such as, traditional machine learning models using TF-IDF, neural networks using pre-trained word embeddings including FastText and GloVe, and state-of-the-art advance transformer based models with advanced contextual embeddings. Additionally we employed a LLM such as GPT-3.5 turbo with advance contextual embeddings. These methods utilized natural language processing (NLP) techniques to develop an advanced automated tool that synthesizes these insights and exposes hidden patterns in the our dataset. The goal was to empower healthcare professionals to effectively monitor withdrawal symptoms and their treatment, enabling personalized care.

This study makes the following contributions:

1. **Manually Annotated Dataset:** We built a multi class and multi-label dataset consisting of 8 different drug classes, each with multiple associated overdose symptoms, enabling predictive modeling and analysis of various drug overdose effects based on user-reported data. This dataset enables in-depth analysis of drug abuse and symptom severity.
2. **Detailed Annotation Process:** We developed a set of detailed annotation guidelines to facilitate accurate labeling of drug names and its overdose effects.

3. **Contextual Understanding of Social Media Discourse:** We addressed key linguistic challenges, such as the use of slang, informal expressions, and context-dependent meanings in drug names during the overdose discussions, ensuring precise labeling of the associated substances.
4. **Contextual Embeddings for Drug and Symptom Detection:** We propose to explore contextual embeddings provided by GPT-4o mini, with fine-tuning to detect specific drug consumed and it overdose effects. This approach aims to assist healthcare professionals, mental health counselors, and policymakers in making informed decisions for better prevention and intervention strategies.
5. **Performance Improvements:** The comprehensive set of experiments demonstrated that the proposed framework achieved benchmark-level performance by outperforming the baselines in both drug abuse and its overdose symptoms. The proposed framework achieved 98% accuracy in drug abuse and 97& in overdose symptoms and showing a 5% and 8%performance improvement in accuracy over the baselines (XGB = 0.90) respectively.

The remainder of this paper is structured as follows: Section II reviews related work and existing literature. Section III describes the methodology and design. Section IV presents the experimental results and their analysis. Section V highlights the limitations of the proposed method. Section VI concludes the paper with a summary of findings and potential directions for future work.

## 2. Literature Review

Zhang et al. [23] proposed a feature selection-based multi-label k-nearest neighbor method (FS-MLKNN) to predict drug side effects by identifying critical feature dimensions and building accurate multi-label models. They further enhanced performance by developing an ensemble learning approach combining individual FS-MLKNN models, outperforming existing methods. Their method improves prediction accuracy while providing interpretable results, making it a promising tool for side effect prediction in drug discovery.

Viera et al. [24] explored the association between alcohol outlets and opioid overdose deaths in Connecticut (2019–2020), finding that one-third of fatalities involved ethanol. They utilized spatial and regression analyses, showing that alcohol outlet density (but not proximity) was linked to ethanol-detected overdoses, even after adjusting for social factors. The results suggest alcohol outlets should be considered in overdose prevention strategies and community interventions.

Zhao et al. [25The authors explored how actin polymerization affects heroin-associated memory reconsolidation in rats using a self-administration model. They found that inhibiting actin polymerization with Latrunculin A in the nucleus accumbens core immediately after memory retrieval reduced heroin-seeking behavior, but delayed treatment had no effect. The results suggest that targeted disruption of memory reconsolidation could offer a potential strategy to prevent relapse in heroin addiction.

Mohd Nazri et al. [26] introduces an EEG-based classification approach for diagnosing alcohol use disorder (AUD) using effective connectivity (EC) features and support vector machine (SVM) algorithms. By analyzing resting-state EEG signals from 35 AUD patients and 35 healthy controls, the method achieved a peak accuracy of 96.37%, demonstrating its potential as an objective diagnostic tool. These findings suggest that EC-based EEG analysis can enhance AUD detection and inform personalized treatment strategies.

Rabinowitz et al. [27] uses random forest to predict early treatment discontinuation in substance use disorder treatment, analyzing data from 29,809 individuals. While the model identified those who stayed, it struggled to predict premature dropout (AUC: 0.631–0.671). Key predictors included treatment center, program type, age, and substance use history, emphasizing facility-level interventions for better retention.

## 3. Methodology

### Construction of dataset

For the construction of our dataset, we collected 1 million Reddit posts from 8 subreddits related to drug use, and their associated overdose symptoms, including r/Drugs, r/Alcoholism, r/MDMA, r/cocaine, r/LSD, r/Opiates, r/OpiatesRecovery, and r/Addiction. Subreddits are individual communities within the Reddit platform where users share their posts in the form of images, videos, or text, and are denoted by the prefix 'r/'. For example, a popular subreddit like r/OpiatesRecovery, allow users to share their struggles and success stories and in r/addiction subreddit where individuals share their advice on addiction and recovery, offering empathy and guidance to those in need. To gather drug overdose data, we first develop a python code and connect the Pushshift.io API that provides access to a massive archive of Reddit data, including posts, comments, and metadata. It was developed as an alternative to Reddit's official API, offering more efficient and flexible data retrieval. Our primary objective was to gather real-life experiences shared by individuals struggling with drug specific drug addiction, overdose effects, and recovery journeys. These posts often cover deeply personal stories—some detailing the first time they used drugs, others describing their battles with overdose. We concentrated on posts mentioning drugs name, and their associated overdose symptoms, ensuring that our dataset captured a wide range of experiences. Reddit allows long posts, with a character limit of 40,000 per post. This makes it popular for in-depth discussions, detailed storytelling, and comprehensive advice. Due to messy and long post it was very difficult to annotate so, we cleaned the data by removing spam, advertisements, and duplicate entries to maintain accuracy. Figure 1 shows the proposed architecture and design of this study.

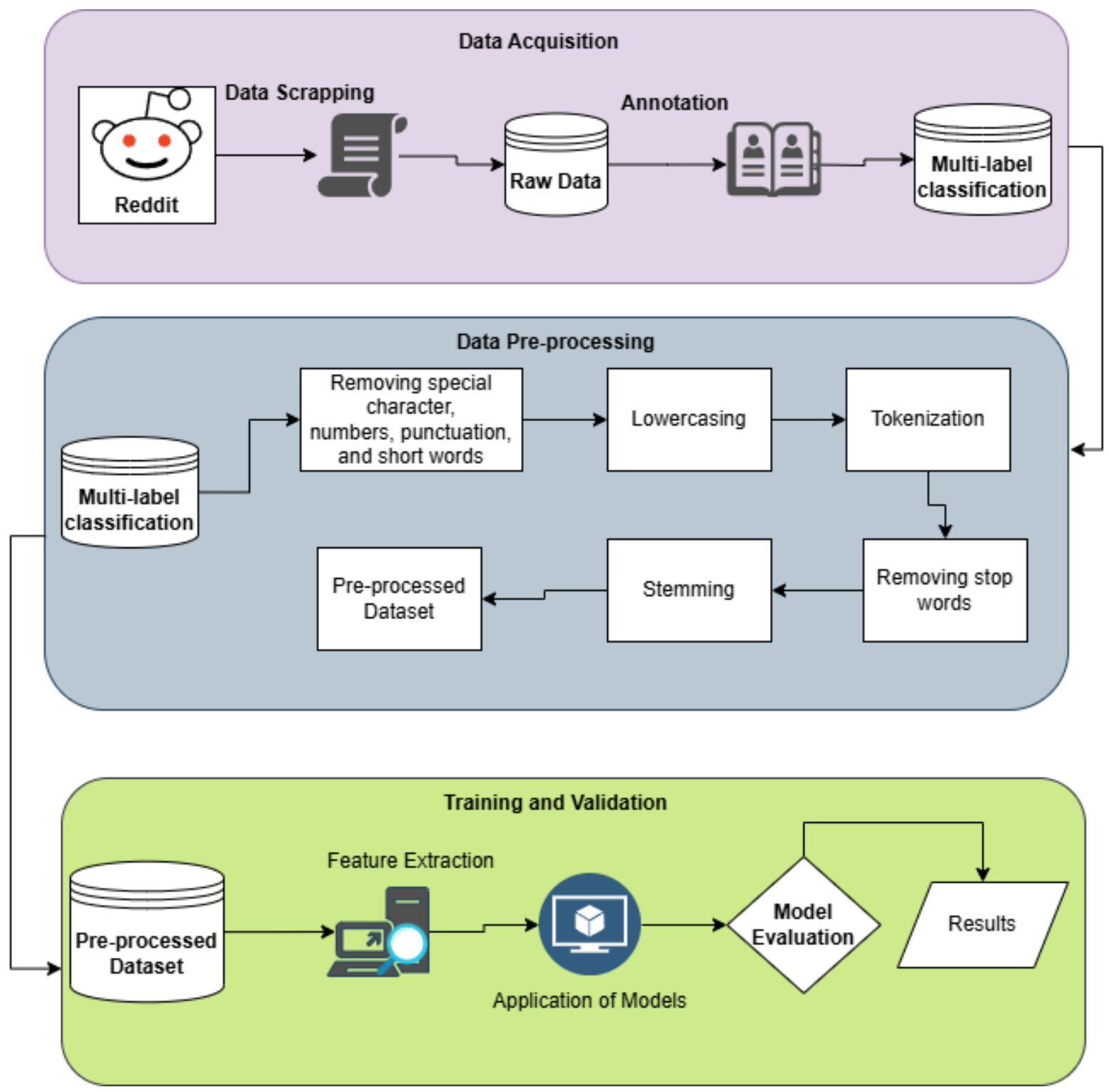

**Figure 1.** Proposed architecture and design.

## Pre-processing

Pre-processing is a crucial phase in machine learning [36-40] that involves cleaning raw data to enable meaningful analysis and improve the performance of machine learning models—especially when dealing with social media data. Social media content often contains multilingual, informal language, emojis, hashtags, and slang, all of which can hinder accurate analysis. To address these challenges, we apply the following pre-processing techniques:

1. Removal of special characters, # tags, Emoji's and digits from posts.
2. Removal of stop words and apply lemmatization and stemming to reduce words to their root forms, enhancing model performance.
3. Removal of duplicate posts and spam, advertisements, and unrelated discussions.
4. We apply filtering techniques to purify only those posts which contains drug names, overdose symptoms, or treatment-related to reverse the overdose effects.
5. We perform normalization to standardize variations of drug names and medical terms, improving consistency.
6. Finally, we balanced the dataset to ensure fair representation of different drug types, overdose effects, and recovery treatments, leading to a well-structured and high-quality dataset for further machine learning analysis.
7. After this rigorous filtering process, only **8,000 posts** were found to be relevant for our study, ensuring that our dataset focused solely on valuable discussions related to drug overdose. This final dataset is proceed further for annotation procedure.

## Annotation

Data annotation is the process of labeling data with predefined categories to make it understandable for machine learning models. It involves tagging raw data such as text, images, or audio. This helps models recognize patterns and make accurate predictions. High-quality annotations are essential for developing effective and reliable supervised learning systems.

## Annotation guidelines

The primary objective of these guidelines is to provide a systematic, reproducible framework for human annotators to manually extract and label drug names, including street names, and their associated overdose symptoms from first-person narrative reports. The dataset consists of unstructured texts detailing drug overdoses, including substances used, and acute symptoms experienced. Human annotators must rely on careful reading, contextual understanding, and toxicological knowledge, using explicit narrative cues and severity indicators to guide labeling, while noting potential biases, such as underreporting of stigmatized substances. These guidelines ensure accurate, high-quality data extraction for public health research and clinical decision support, upholding methodological integrity and alignment with toxicological standards.

To identify drug names, human annotators must extract the pharmacologically active substance explicitly cited as the primary cause of the overdose event, which may appear as clinical names (e.g., "Fentanyl," "Xanax"), brand names (e.g., "Valium"), or street names (e.g., "smack" for heroin, "bars" for Xanax, "china white" for fentanyl). Annotators should record the drug exactly as it appears in the narrative to preserve fidelity, but for clarity, they must manually map street names to their clinical equivalents (e.g., "molly" to "MDMA") using a reference list of common drug slang, such as those compiled by public health agencies or toxicology resources. In cases of polydrug use (e.g., "I took molly and dope"), annotators should prioritize the drug most directly linked to the overdose, guided by clear narrative cues like "overdosed on dope" (mapped to "heroin"), phrases such as "dope hit me hardest," causal words like "because of," or repeated mentions of one drug. If attribution remains unclear, symptoms are assigned to all mentioned drugs, and annotators should add a note labeled "polydrug uncertainty" to flag the ambiguity for further review. Vague terms like "stuff" or "pills" are excluded unless the narrative specifies their identity (e.g., "stuff was heroin"), and unrelated substances (e.g., "weed" in an opioid-focused narrative) are disregarded unless causally implicated. To ensure accuracy, annotators must cross-check drug names, including street names, against a reliable source like a toxicology textbook or drug slang glossary, correcting errors (e.g., "H" to "Heroin") and documenting any uncertainties.

Overdose symptoms are defined as acute, drug-induced physiological, psychological, or behavioral manifestations reported in the narrative, indicating a toxicological emergency. Human annotators should carefully identify symptoms explicitly tied to the overdose event, such as "nausea," "bluish lips," or "slow breathing," allowing multiple symptoms per drug to reflect the complexity of overdoses (e.g., "Fentanyl" causing "respiratory depression," "drowsiness," "vomiting"). Symptoms must be consistent with known overdose effects—opioids typically cause respiratory depression and sedation, stimulants lead to tachycardia and agitation, hallucinogens induce confusion or hallucinations, and depressants result in drowsiness or cyanosis. Chronic conditions (e.g., "liver pain") or emotional states (e.g., "guilt") are excluded unless clearly triggered by the overdose. Annotators should confirm symptoms are linked to drug use through timing (e.g., soon after taking the drug) or causal statements (e.g., “Fentanyl made me pass out”), using severity clues like “nearly died” or “needed Narcan” to verify overdose context. The narrative’s wording (e.g., “blue lips”) should be recorded as is, but annotators may note clinical terms (e.g., “cyanosis”) in parentheses for clarity, ensuring the original text is preserved. To distinguish overdose from withdrawal symptoms (e.g., “shakes” after stopping alcohol), annotators should assess whether symptoms occurred shortly after drug use (suggesting overdose) or after a clear pause, like “the next day” (suggesting withdrawal). For ambiguous cases (e.g., “Alcohol caused shaking after a binge”), annotators should use a checklist: look for acute toxicity signs (e.g., “Narcan used”), check if symptoms match overdose patterns, and note any mention of stopping the drug. Unclear cases should be flagged with detailed notes explaining the reasoning. Effects from medical interventions (e.g., “taste of charcoal”) are not counted as symptoms. Table 1 presents examples from the labeled overdose dataset, showcasing drug use along with the associated overdose symptoms. Each entry reflects a real-world account of drug use and its acute physiological effects, extracted and categorized for analysis.

Table 1: Drug Use and Associated Overdose Symptoms from the Dataset.

| Extracted Post | Drug Name | Overdose Effects |
|---|---|---|
| "Had a few drinks with some friends at a bar. Took a shot of whiskey, then another, and started feeling really dizzy. I couldn’t stand up and my vision went blurry. I passed out right there. Woke up with a massive headache and in the hospital." | Alcohol (whiskey) | Dizziness, fainting, blurred vision, headache |
| "Tried cocaine for the first time at a party. I snorted a line, and for the next 20 minutes, I felt super energized. But soon after, my heart was racing, and I couldn’t breathe. I started shaking uncontrollably. My friends had to take me to the ER." | Cocaine | Tachycardia (increased heart rate), shortness of breath, tremors |
| "I took some ecstasy at a rave. Felt great at first, but after a while, I started feeling really hot and dizzy. My heart was pounding and I couldn’t stop sweating. I had to sit down, and I felt like I was going to pass out. My friends took me outside to cool off." | Ecstasy (MDMA) | Hyperthermia (high body temperature), tachycardia, sweating, dizziness |
| "I didn't know the pills I took were laced with fentanyl. I felt really tired and dizzy, and I couldn't keep my eyes open. Then, I couldn’t breathe, and everything went black. I woke up in the ER with Narcan administered to me." | Fentanyl | Respiratory depression, drowsiness, loss of consciousness |
| "I injected heroin for the first time. I immediately felt the rush, but then my breathing slowed down. I started to feel nauseous, and next thing I know, I was out cold. I woke up with paramedics around me after they revived me." | Heroin (smack) | Respiratory depression, nausea, coma |
| "I took a couple hits of ketamine at a party. Everything started to feel distant and strange. I couldn’t walk, and I felt like I was floating. I eventually blacked out and woke up in the hospital." | Ketamine | Dissociation, confusion, loss of consciousness |
| "Took some acid at a camping trip. I felt good at first, but then everything started to warp. Colors looked weird, and I heard voices. I couldn’t tell what was real anymore. I panicked and had to be restrained by my friends." | LSD (acid) | Visual and auditory hallucinations, confusion, paranoia |

| "I smoked meth with a friend late at night. After a few hits, I felt like I was on top of the world, but then I started feeling really agitated. My heart was racing, and I couldn't stop moving. I felt like I was overheating and had to cool off outside." | Methamphetamine (meth) | Tachycardia, agitation, hyperthermia, restlessness |
|---|---|---|

## Annotation Selection

To ensure the accuracy and reliability of the annotation procedure, especially for classifying specific drugs abuse by the user and their associated overdose symptoms we employed three annotators as they were Ph.D. students in Computer Science with expertise in machine learning, deep learning, and large language models (LLMs) and have enough knowledge of data annotation[45] [46]. Additionally, they were native English speakers, which contributed to a more precise understanding of linguistic nuances commonly found in the dataset.

## Annotation procedure

To construct high-quality dataset for drug abuse detection and overdose symptom classification, we adopted a hybrid data annotation approach combining expert human annotation with large language model (LLM)-assisted labeling. Initially, we manually labeled 2,000 samples related to drug abuse, and its associated overdose symptoms, with special focus on handling linguistic challenges such as slang, informal expressions, and context-dependent meanings to ensure precise labeling. As users often use slang terms to refer to various substances, and recognizing these terms is crucial for accurate detection in drug detection and its associated overdose symptoms. For example, **alcohol** is commonly referred to as booze, liquor, brew, sauce, or hooch. **Heroin** has many slang names, including smack, dope, junk, black tar, horse, and China White. **Cocaine** is frequently called coke, blow, snow, powder, yayo, or white. **Ecstasy (MDMA)** is often referred to as molly, E, X, adam, or ecstasy. This dataset was then used to fine-tune GPT-3.5 Turbo, which was employed to annotate a 2000 more samples. After the model classification three human carefully review and correct the incorrect or inconsistent labels, resulting in 4,000 accurately labeled samples. These were used to retrain the LLM model, which was then utilized to annotate an additional 1,114 samples, followed by another round of manual review and correction. This iterative semi-supervised approach yielded a final dataset of 5,114 samples, each annotated for multi-label and multi-class classification tasks with high accuracy and consistency.

## Inter-Annotator agreement

To assess the quality and consistency of annotations in our multi-class and multi-label dataset, we used Fleiss' Kappa as the inter-annotator agreement metric. Each instance in the dataset was first labeled with the identified drug name, followed by the associated symptoms in a multi-label format. Three annotators independently labeled the dataset. The calculated Kappa value for drug identification was 0.83, while the Kappa value for overdose symptom annotation was 0.79. These scores indicate a high level of consistency between annotators, supporting the robustness of the annotation framework. Table 2 shows the Interpretation of the kappa values.

Table 2. Interpretation of the kappa values.

| Kappa Value Range | Interpretation |
|---|---|
| 1.0 | Perfect agreement |
| 0.80 to 1.0 | Substantial agreement |
| 0.60 to 0.80 | Moderate agreement |
| 0.40 to 0.60 | Fair agreement |
| <40 | Poor agreement |

## Application of models

In this section we will discuss the overall workflow of our predictive modeling framework for identifying drug types and associated overdose symptoms from Reddit posts. Starting with the final annotated dataset, we split the data into a learning set (80%) and a testing set (20). The learning set is then used for training models through Machine Learning such as SVM, LR, DT and XGBoost, Deep Learning such as CNN, BiLSTM, and Transfer Learning such as Bio-BERT, Clinical BERT and XLM-roberta techniques. For Machine learning, we utilized TF-IDF vectorization with traditional classifiers; for Deep learning, we employed pretrained FastText and GloVe embeddings with neural networks; and for TL, we fine-tuned transformer-based models to leverage advanced contextual embeddings. The trained models then make predictions on the unseen testing set. These predictions are evaluated and compared to the actual labels to produce the final predicted values. The framework integrates linguistic nuances in user-generated content and enables multi-label classification for complex symptom detection. This pipeline ensures robust performance across multiple modeling paradigms. To assess the performance of our models we utilized four standard evaluation metrics such as Precision, Recall, F1-Score, and Accuracy, as shown in Equations (1), (2), (3) and (4). Figure 2 show the overall work flow of our proposed models.

$$\text{Accuracy} = \frac{\text{TN+TP}}{\text{Total Predictions}} \tag{1}$$

$$\text{Percision} = \frac{\text{TP}}{\text{FP+TP}} \tag{2}$$

$$\text{Recall} = \frac{\text{TP}}{\text{FN+TP}} \tag{3}$$

$$\text{F1} - \text{score} = 2 \times \frac{\text{Recall} \times \text{Precision}}{\text{Recall+ Precision}} \tag{4}$$

While TP is true positive, FP is false positive and TN is true negative.

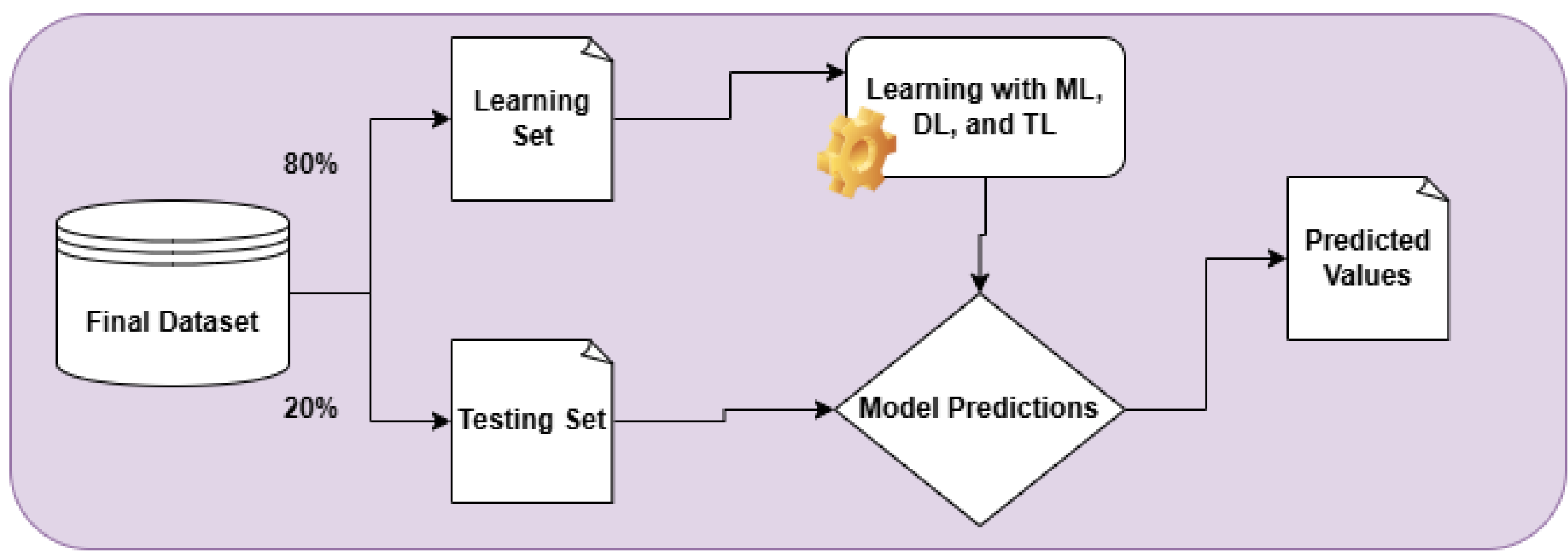

Figure 2. Application of models, training and testing phase.

## 4. Results and Analysis

### Machine learning Results

Table 3 compares the performance of five machine learning models—Logistic Regression (LR), Decision Tree (DT), Random Forest (RF), K-Nearest Neighbors (KNN), and Naive Bayes (NB)—on two classification tasks: identifying Drug Names and recognizing Overdose Effects. For each task, the models are evaluated using four metrics: Precision, Recall, F1-Score, and Accuracy. In the Drug Name classification, LR achieves the best performance across all metrics (0.92), closely followed by DT (0.91) and RF (0.89). NB performs moderately well (0.77), while KNN significantly underperforms (0.42). For the Overdose Effects classification, RF yields the highest scores (0.91), slightly outperforming DT (0.90) and LR (reported vaguely as "0.89," presumably meant to be 0.89). Again, NB and KNN lag behind with scores of 0.65 and 0.67, respectively. Overall, LR, DT, and RF consistently show strong and reliable performance, whereas KNN is the weakest model across both tasks, and NB shows moderate capability.

Table 3. Results for machine learning models.

| Class | Model | Precision | Recall | F1-Score | Accuracy |
|---|---|---|---|---|---|
| Drug Name | LR | 0.92 | 0.92 | 0.92 | 0.92 |
| | DT | 0.91 | 0.91 | 0.91 | 0.91 |
| | RF | 0.89 | 0.89 | 0.89 | 0.89 |
| | KNN | 0.42 | 0.42 | 0.42 | 0.42 |
| | NB | 0.77 | 0.77 | 0.77 | 0.77 |
| Overdose Effects | LR | 0..89 | 0..89 | 0..89 | 0..89 |
| | DT | 0.9 | 0.9 | 0.9 | 0.9 |
| | RF | 0.91 | 0.91 | 0.91 | 0.91 |
| | KNN | 0.67 | 0.67 | 0.67 | 0.67 |
| | NB | 0.65 | 0.65 | 0.65 | 0.65 |

## Deep learning Results

Table 4 presents the evaluation results of two deep learning models—Convolutional Neural Network (CNN) and Bidirectional Long Short-Term Memory (BiLSTM)—on the classification tasks of Drug Name identification and Overdose Effects detection. The performance is measured using four metrics: Precision, Recall, F1-Score, and Accuracy. In the Drug Name classification, CNN outperforms BiLSTM with consistently higher scores (0.94 across all metrics compared to BiLSTM's 0.89), indicating superior ability to correctly identify drug names. Similarly, in the Overdose Effects classification, CNN again demonstrates stronger performance with values of 0.91, while BiLSTM achieves only 0.74 across all metrics, suggesting a relatively weaker capability in capturing the nuances of overdose-related textual features. Overall, CNN proves to be the more effective model in both tasks, offering higher reliability and accuracy than BiLSTM.

Table 4. Results for Deep learning models

| Class | Model | Precision | Recall | F1-Score | Accuracy |
|---|---|---|---|---|---|
| Drug name | CNN | 0.94 | 0.94 | 0.94 | 0.94 |
| | BiLSTM | 0.89 | 0.89 | 0.89 | 0.89 |
| Overdose Effects | CNN | 0.91 | 0.91 | 0.91 | 0.91 |
| | BiLSTM | 0.74 | 0.74 | 0.74 | 0.74 |

## Transformer Results

The table 5 shows the performance of three language-based transformer models such as BERT, Bio-BERT, and XLM-R on multi-label and multi-class classification tasks including "Drug Name" and "Overdose Effects." Each model's performance is assessed using four metrics such Precision, Recall, F1-Score, and Accuracy. For the "Drug Name" classification, BERT model attains the maximum scores across all metrics (0.94), slightly outperforming Bio-BERT (0.93) and XLM-R (0.91), representing its superior performance in identifying drug names. On the other hand, for the "Overdose Effects" which is multi-label task, XLM-R performs the top with a consistent score =0.94, followed by Bio-BERT (0.92) and BERT (0.90), suggesting that XLM-R's multilingual and diverse pre-training enables it to better capture the hidden pattern in overdose effects. Overall, while BERT excels in classifying drug names, while XLM-R shows more effective in detecting overdose-related effects.

Table 5. Results for Transformer.

| Class | Model | Precision | Recall | F1-Score | Accuracy |
|---|---|---|---|---|---|
| Drug Name | BERT | 0.94 | 0.94 | 0.94 | 0.94 |
| | Bio-BERT | 0.93 | 0.93 | 0.93 | 0.93 |
| | XLM-R | 0.99 | 0.99 | 0.99 | 0.99 |
| Overdose Effects | BERT | 0.9 | 0.9 | 0.9 | 0.9 |
| | Bio-BERT | 0.92 | 0.92 | 0.92 | 0.92 |
| | XLM-R | 0.99 | 0.99 | 0.99 | 0.99 |

## Large Language models Results

The figure 3 presents the performance metrics of a Language Model such as GPT-4o-mini across multiple evaluation runs, showing consistently high scores in **Precision (0.98)**, **Recall (0.97-0.98)**, **F1-Score (0.97)**, and **Accuracy (0.96-0.98)**. These near-perfect numbers suggest the model is highly reliable—like a sharp-eyed proofreader catching almost every mistake while rarely mislabeling anything. The tight clustering of scores across runs indicates stability, meaning you can trust its predictions as much as you'd trust a seasoned expert making careful, repeatable judgments. Whether detecting drug interactions or analyzing medical text, this model performs with impressive consistency, offering both precision and dependability.

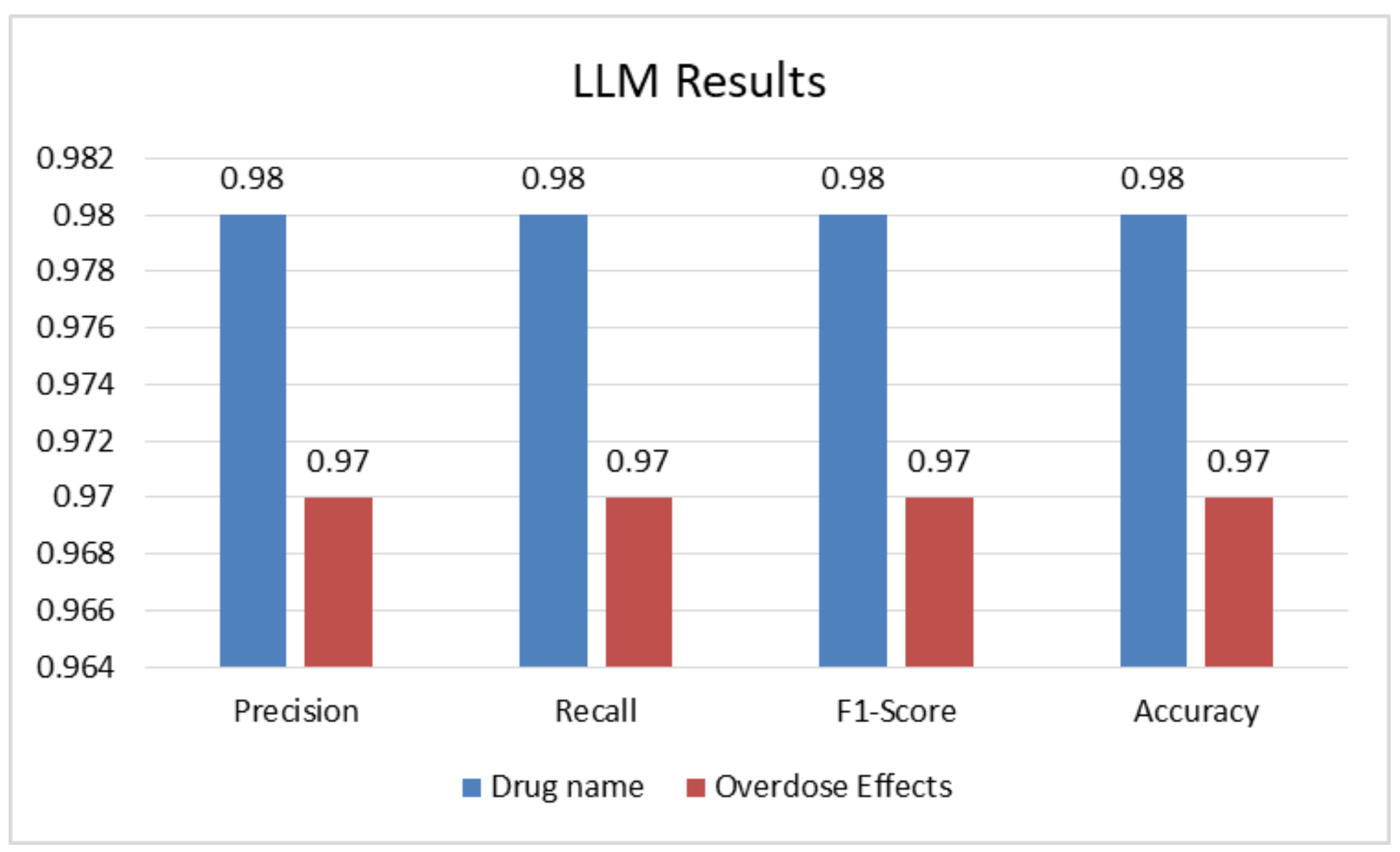

Figure 3. Results for Large language models.

References

1. Islam, M. S., Sarkar, T., Khan, S. H., Kamal, A. H. M., Hasan, S. M., Kabir, A., ... & Seale, H. (2020). COVID-19–related infodemic and its impact on public health: A global social media analysis. *The American journal of tropical medicine and hygiene*, *103*(4), 1621.
2. Kim, S. J., Marsch, L. A., Hancock, J. T., & Das, A. K. (2017). Scaling up research on drug abuse and addiction through social media big data. *Journal of medical Internet research*, *19*(10), e353.
3. Rizzo, A., Calandi, L., Faranda, M., Rosano, M. G., Carlotta, V., & Vinci, E. (2025). Stigma against mental illness and mental health: The role of Social Media. *Adv Med Psychol Public Health*, *2*(2), 125-130.
4. McGorry, P., Gunasiri, H., Mei, C., Rice, S., & Gao, C. X. (2025). The youth mental health crisis: analysis and solutions. *Frontiers in Psychiatry*, *15*, 1517533.

5. Lokala, U., Phukan, O. C., Dastidar, T. G., Lamy, F., Daniulaityte, R., & Sheth, A. (2024). Detecting substance use disorder using social media data and the dark web: time-and knowledge-aware study. *JMIRx Med*, *5*, e48519.
6. Michael, W. Usage of Social Media as a Tool for Eradicating Drug Abuse among University Undergraduates in Nigeria.
7. Gustafsson, M., Silva, V., Valeiro, C., Joaquim, J., van Hunsel, F., & Matos, C. (2024). Misuse, Abuse and Medication Errors' Adverse Events Associated with Opioids—A Systematic Review. *Pharmaceuticals*, *17*(8), 1009.
8. Oksanen, A., Miller, B. L., Savolainen, I., Sirola, A., Demant, J., Kaakinen, M., & Zych, I. (2021). Social media and access to drugs online: A nationwide study in the United States and Spain among adolescents and young adults. *European journal of psychology applied to legal context*, *13*(1), 29-36.
9. Zhao, J., Jia, T., Wang, X., Xiao, Y., & Wu, X. (2022). Risk factors associated with social media addiction: An exploratory study. *Frontiers in Psychology*, *13*, 837766.
10. Taylor, S., Paluszek, M. M., Rachor, G. S., McKay, D., & Asmundson, G. J. (2021). Substance use and abuse, COVID-19-related distress, and disregard for social distancing: A network analysis. *Addictive behaviors*, *114*, 106754.
11. Gorfinkel, L., Stohl, M., Shmulewitz, D., & Hasin, D. (2024). Self-reported substance use with clinician interviewers versus self-administered surveys. *Journal of Studies on Alcohol and Drugs*, *85*(1), 92-99.
12. Kemble, Hannah, et al. "Children and young people's self-reported experiences of asthma and self-management nursing strategies: an integrative review." *Journal of Pediatric Nursing* 77 (2024): 212-235.
13. Brose, L. S., Reid, J. L., Robson, D., McNeill, A., & Hammond, D. (2024). Associations between vaping and self-reported respiratory symptoms in young people in Canada, England and the US. *BMC medicine*, *22*(1), 213.
14. Olalekan Kehinde, A. (2025). Leveraging Machine Learning for Predictive Models in Healthcare to Enhance Patient Outcome Management. *Int Res J Mod Eng Technol Sci*, *7*(1), 1465.
15. Kazi, K. S. L., & Mahant, M. A. (2025). Machine Learning-Driven Internet of Things (MLIoT)-Based Healthcare Monitoring System. In *Digitalization and the Transformation of the Healthcare Sector* (pp. 205-236). IGI Global Scientific Publishing.
16. Liyakat, K. K. S. (2025). Heart health monitoring using IoT and machine learning methods. In *AI-Powered Advances in Pharmacology* (pp. 257-282). IGI Global.
17. Bernstorff, M., Hansen, L., Enevoldsen, K., Damgaard, J., Hæstrup, F., Perfalk, E., ... & Østergaard, S. D. (2025). Development and validation of a machine learning model for prediction of type 2 diabetes in patients with mental illness. *Acta Psychiatrica Scandinavica*, *151*(3), 245-258.
18. Delgadillo, J., & Lutz, W. (2026). Precision mental health care for depression. *APA handbook of depression. American Psychological Association*.
19. Maurya, R. K., Montesinos, S., Bogomaz, M., & DeDiego, A. C. (2025). Assessing the use of ChatGPT as a psychoeducational tool for mental health practice. *Counselling and Psychotherapy Research*, *25*(1), e12759.

20. Roos, C. R., Kober, H., Trull, T. J., MacLean, R. R., & Mun, C. J. (2020). Intensive longitudinal methods for studying the role of self-regulation strategies in substance use behavior change. *Current addiction reports*, *7*, 301-316.
21. Vázquez, A. L., Domenech Rodríguez, M. M., Barrett, T. S., Schwartz, S., Amador Buenabad, N. G., Bustos Gamiño, M. N., ... & Villatoro Velázquez, J. A. (2020). Innovative identification of substance use predictors: machine learning in a national sample of Mexican children. *Prevention Science*, *21*, 171-181.
22. Kar, S. B., & Alex, S. (2020). Public health approaches to substance abuse prevention: A multicultural perspective. In *Substance Abuse Prevention* (pp. 11-41). Routledge.
23. Zhang, W., Liu, F., Luo, L., & Zhang, J. (2015). Predicting drug side effects by multi-label learning and ensemble learning. *BMC bioinformatics*, *16*, 1-11.
24. Viera, A., Heimer, R., & Grau, L. E. (2025). Spatial associations between alcohol detection in opioid overdose deaths and alcohol outlets. *Drug and Alcohol Dependence*, 112659.
25. Zhao, H., Li, H., Meng, L., Du, P., Mo, X., Gong, M., ... & Liao, Y. (2025). Disrupting heroin-associated memory reconsolidation through actin polymerization inhibition in the nucleus accumbens core. *International Journal of Neuropsychopharmacology*, *28*(1), pyae065.
26. Mohd Nazri, A. K., Yahya, N., Khan, D. M., Mohd Radzi, N. I. Z., Badruddin, N., Abdul Latiff, A. H., & Abdulaal, M. J. (2025). Partial directed coherence analysis of resting-state EEG signals for alcohol use disorder detection using machine learning. *Frontiers in Neuroscience*, *18*, 1524513.
27. Rabinowitz, J. A., Wells, J. L., Kahn, G., Ellis, J. D., Strickland, J. C., Hochheimer, M., & Huhn, A. S. (2025). Predictors of treatment attrition among individuals in substance use disorder treatment: A machine learning approach. *Addictive Behaviors*, 108265.
28. Bohnert, A. S., Ilgen, M. A., Ignacio, R. V., McCarthy, J. F., Valenstein, M., & Blow, F. C. (2012). Risk of death from accidental overdose associated with psychiatric and substance use disorders. *American Journal of Psychiatry*, *169*(1), 64-70.
29. Haegele, J. A., Ross-Cypcar, S. M., & Garcia, J. M. (2025). Illicit drug use among adolescents and young adults with impairments in the US: A cross-sectional analysis of the National Survey on Drug Use And Health. *Preventive Medicine*, *191*, 108222.
30. Ahmad, M., Farid, H., Ameer, I., Muzamil, M., Jalal, A. H. M., Batyrshin, I., & Sidorov, G. (2025). Opioid Named Entity Recognition (ONER-2025) from Reddit. *arXiv preprint arXiv:2504.00027*.
31. Alnawas, A., Alharbi, H., & Al-Jawad, M. M. H. (2025). Understanding the Influence Impact of Social Media on Drug Addiction: A Novel Sentiment Analysis Approach Using Multi-Level User Engagement Data.
32. Ma, L., Chen, R., Ge, W., Rogers, P., Lyn-Cook, B., Hong, H., ... & Zou, W. (2025). AI-powered topic modeling: comparing LDA and BERTopic in analyzing opioid-related cardiovascular risks in women. *Experimental Biology and Medicine*, *250*, 10389.
33. Kim, S., Lee, H., Woo, S., Lee, H., Park, J., Kim, T., ... & Yon, D. K. (2025). Global, regional, and national trends in drug use disorder mortality rates across 73 countries from 1990 to 2021, with projections up to 2040: a global time-series analysis and modelling study. *EClinicalMedicine*, *79*.
34. National Institute on Drug Abuse. (2023). *Fiscal Year 2024 Budget Information - Congressional Justification for National Institute on Drugs and Addiction*. National Institutes of Health. https://nida.nih.gov/about-nida/legislative-activities/budget-information/fiscal-year-2024-budget-information-congressional-justification-national-institute-drug-abuse.
35. Hedegaard, H., Bastian, B. A., Trinidad, J. P., Spencer, M., & Warner, M. (2018). Drugs most frequently involved in drug overdose deaths: United States, 2011–2016.

36. Ahmed, M., Usman, S., Shah, N. A., Ashraf, M. U., Alghamdi, A. M., Bahadded, A. A., & Almarhabi, K. A. (2022). AAQAL: A machine learning-based tool for performance optimization of parallel SPMV computations using block CSR. *Applied Sciences*, *12*(14), 7073.
37. Ullah, F., Zamir, M., Arif, M., Ahmad, M., Felipe-Riveron, E., & Gelbukh, A. (2024, March). Fida@ dravidianlangtech 2024: A novel approach to hate speech detection using distilbert-base-multilingual-cased. In *Proceedings of the Fourth Workshop on Speech, Vision, and Language Technologies for Dravidian Languages* (pp. 85-90).
38. Ullah, F., Zamir, M. T., Ahmad, M., Sidorov, G., & Gelbukh, A. (2024). Hope: A multilingual approach to identifying positive communication in social media. In *Proceedings of the Iberian Languages Evaluation Forum (IberLEF 2024), co-located with the 40th Conference of the Spanish Society for Natural Language Processing (SEPLN 2024), CEUR-WS. org*.
39. Ahmad, M., Sardar, U., Batyrshin, I., Hasnain, M., Sajid, K., & Sidorov, G. (2024). Elegante: A Machine Learning-Based Threads Configuration Tool for SpMV Computations on Shared Memory Architecture. *Information*, *15*(11), 685.
40. Ahmad, M., Usman, S., Farid, H., Ameer, I., Muzammil, M., Hamza, A., ... & Batyrshin, I. (2024). Hope Speech Detection Using Social Media Discourse (Posi-Vox-2024): A Transfer Learning Approach. *Journal of Language and Education*, *10*(4 (40)), 31-43.
41. Usman, Muhammad, et al. "Fine-Tuned RoBERTa Model for Bug Detection in Mobile Games: A Comprehensive Approach." *Computers* 14.4 (2025): 113.
42. Ahmad, Muhammad, et al. "Cotton Leaf Disease Detection Using Vision Transformers: A Deep Learning Approach." *crops* 1 (2024): 3.
43. Ahmad, M., Farid, H., Ameer, I., Muzamil, M., Jalal, A. H. M., Batyrshin, I., & Sidorov, G. (2025). Opioid Named Entity Recognition (ONER-2025) from Reddit. *arXiv preprint arXiv:2504.00027*.
44. Ahmad, M., Ameer, I., Sharif, W., Usman, S., Muzamil, M., Hamza, A., ... & Sidorov, G. (2025). Multilingual hope speech detection from tweets using transfer learning models. *Scientific Reports*, *15*(1), 9005.
45. Ahmad, M., Sidorov, G., Amjad, M., Ameer, I., & Batyrshin, I. (2025). Opioid Crisis Detection in Social Media Discourse Using Deep Learning Approach. *Information*, *16*(7), 545.
46. Ahmad, M., Batyrshin, I., & Sidorov, G. (2025). Sentiment Analysis Using a Large Language Model–Based Approach to Detect Opioids Mixed With Other Substances Via Social Media: Method Development and Validation. *JMIR infodemiology*, *5*, e70525.
47. Ahmad, M., Waqas, M., Hamza, A., Usman, S., Batyrshin, I., & Sidorov, G. (2025). UA-HSD-2025: Multi-Lingual Hate Speech Detection from Tweets Using Pre-Trained Transformers. *Computers*, *14*(6), 239.
48. Ahmad, M., H. Farid, I. Ameer, F. Ullah, M. Muzamil, M. Jalal, A. Hamza, I. Batyrshin, and G. Sidorov. "UE-NER-2025: A GPT-based Approach to Multilingual Named Entity Recognition on Urdu and English." *IEEE Access* (2025).
49. Ahmad, Muhammad, Muhammad Waqas, Ameer Hamza, Ildar Batyrshin, and Grigori Sidorov. "Hope Speech Detection in code-mixed Roman Urdu tweets: A Positive Turn in Natural Language Processing." *arXiv preprint arXiv:2506.21583* (2025).